\documentclass[sigconf]{acmart}
\AtBeginDocument{%
  }

 \copyrightyear{2025} 
\acmYear{2025} 
\setcopyright{cc}
 \setcctype{by}
 \acmConference[CIKM '25]{Proceedings of the 34th ACM International
 Conference on Information and Knowledge Management}{November 10--14,
2025}{Seoul, Republic of Korea}
 \acmBooktitle{Proceedings of the 34th ACM International Conference on
 Information and Knowledge Management (CIKM '25), November 10--14, 2025,
 Seoul, Republic of Korea}\acmDOI{10.1145/3746252.3761261}
 \acmISBN{979-8-4007-2040-6/2025/11}

\begin{document}

\title{FinCast: A Foundation Model for Financial Time-Series Forecasting}


\author{Zhuohang Zhu}
\email{zzhu6520@uni.sydney.edu.au}
\orcid{0009-0009-8362-281X}
\affiliation{%
  \department{School of Computer Science}
  \institution{The University of Sydney}
  \city{Sydney}
  \state{NSW}
  \country{Australia}
}

\author{Haodong Chen}
\email{haodong.chen@sydney.edu.au}
\orcid{0000-0003-2254-5629}
\affiliation{%
  \department{School of Computer Science}
  \institution{The University of Sydney}
  \city{Sydney}
  \state{NSW}
  \country{Australia}
}

\author{Qiang Qu}
\email{vincent.qu@sydney.edu.au}
\orcid{0000-0002-6648-5050}
\affiliation{%
  \department{School of Computer Science}
  \institution{The University of Sydney}
  \city{Sydney}
  \state{NSW}
  \country{Australia}
}

\author{Vera Chung}
\email{vera.chung@sydney.edu.au}
\orcid{0000-0002-3158-9650}
\affiliation{%
  \department{School of Computer Science}
  \institution{The University of Sydney}
  \city{Sydney}
  \state{NSW}
  \country{Australia}
}


\renewcommand{\shortauthors}{Zhuohang Zhu et al.}

\begin{abstract}

Financial time-series forecasting is critical for maintaining economic stability, guiding informed policymaking, and promoting sustainable investment practices. However, it remains challenging due to various underlying pattern shifts. These shifts arise primarily from three sources: temporal non-stationarity (distribution changes over time), multi-domain diversity (distinct patterns across financial domains such as stocks, commodities, and futures), and varying temporal resolutions (patterns differing across per-second, hourly, daily, or weekly indicators). While recent deep learning methods attempt to address these complexities, they frequently suffer from overfitting and typically require extensive domain-specific fine-tuning. To overcome these limitations, we introduce \textbf{FinCast}, the first foundation model specifically designed for financial time-series forecasting, trained on large-scale financial datasets. Remarkably, \textbf{FinCast} exhibits robust zero-shot performance, effectively capturing diverse patterns without domain-specific fine-tuning. Comprehensive empirical and qualitative evaluations demonstrate that \textbf{FinCast} surpasses existing state-of-the-art methods, highlighting its strong generalization capabilities.

\end{abstract}


  

\maketitle

\section{Introduction}
Forecasting financial time series is crucial for supporting economic stability, guiding investment decisions~\cite{tsay2005analysis}, and managing financial risk~\cite{chatfield2013analysis}. Reliable forecasts help allocate capital efficiently, reduce exposure to market shocks, and inform regulatory policy~\cite{taylor2011asset}. From central banks setting interest rates to institutional investors managing portfolios, accurate forecasts enable timely, data-driven decisions that influence both short-term market movements~\cite{franses1998time} and long-term economic outcomes~\cite{cochrane1997time}.

Despite its importance, financial time-series forecasting remains highly challenging due to various underlying pattern shifts~\cite{ACL_18, tsay2005analysis}. 
First, financial time series are inherently \emph{non-stationary}~\cite{taylor2011asset}: their distribution shifts over time due to factors such as structural economic changes, shifting investor behavior, policy interventions, and technological disruptions. For example, the distribution of prices for a stock like Apple differs significantly between 2021 and 2025, shaped by both macroeconomic conditions and firm-level developments.
Second, forecasting across financial domains poses a core modeling challenge. Each domain, such as stocks, commodities, or currencies, exhibits distinct patterns shaped by diverse factors such as economic mechanisms, regulatory environments, and market structures~\cite{franses1998time}.
Third, financial time series occur at varying temporal resolutions, from second-level tick data to weekly or monthly indicators~\cite{cochrane1997time}. For example, high-frequency data reflect rapid, noise-driven fluctuations, while lower-frequency data capture slower, macro-driven trends. These dynamics are often incompatible, and models designed for a single resolution typically fail to generalize across different temporal resolutions~\cite{chatfield2013analysis}.

%

Existing forecasting models often fail under real-world conditions: they struggle to generalize across distribution shifts, financial domains, and temporal resolutions. Models trained on one financial domain or temporal resolution typically perform poorly when applied elsewhere, and their accuracy degrades rapidly when distributional properties shift. A central reason is that most approaches rely on supervised learning with strong assumptions about the stability of underlying patterns. Their architectures are often tailored to specific financial domains (e.g., stocks) or temporal resolutions (e.g., daily data), which limits their applicability beyond the original setting. Trained on fixed datasets and optimized for narrow tasks, these models overfit to historical patterns and are unable to generalize beyond their original context.

To address the limitations of existing approaches, we introduce \textbf{FinCast}, a foundation model for financial time-series forecasting. Intuitively, a large-capacity model, trained on sufficiently diverse and large-scale financial data, can learn a broad spectrum of temporal patterns, domain-specific dynamics, and resolution-dependent behaviors. FinCast is implemented as a large decoder-only transformer and trained on over 20 billion time points across a wide range of financial domains and temporal resolutions.
To enable this generalization in practice, we introduce three key design choices.
First, Point-Quantile loss (PQ-loss), which jointly optimizes point forecasts and quantile-based probabilistic estimates to model uncertainty across the distribution, enhances robustness to temporal shifts and prevents forecast collapse.
Second, a token-level sparse Mixture-of-Experts (MoE) mechanism that increases capacity efficiently and enables experts to specialize across domains.
Third, learnable frequency embeddings that encode temporal characteristics at varying resolutions, improving the capture of cyclic and seasonal patterns. 
This unified framework balances high capacity with robustness, allowing FinCast to learn both shared and domain-specific dynamics across financial time series.



Empirical evaluations validate the effectiveness of our approach. FinCast consistently outperforms state-of-the-art methods across both zero-shot and supervised financial forecasting benchmarks, achieving best results without task-specific fine-tuning. Our experiments span a wide range of financial domains, including stocks, cryptocurrencies, forex, and futures, capturing the diversity and non-stationarity of real-world markets. Complementary qualitative analyses further show that FinCast adapts well to shifting patterns across domains and temporal resolutions.
Our contributions can be summarised as follows:
\begin{itemize}
    \item We introduce the first foundation model for 
    financial time-series forecasting, a decoder-only transformer with 1B parameters, trained on 20B+ time points across diverse financial domains and temporal resolutions.
    \item We propose a novel Point-Quantile Loss that combines point forecasts with quantile-based probabilistic estimates to enhance robustness under temporal non-stationarity.
    \item We design a learnable frequency embedding that encodes temporal resolution, enhancing adaptability across various temporal resolutions. Combined with a token-level sparse Mixture-of-Experts mechanism, this increases model's capacity efficiently and enables expert specialization across financial domains. 
    \item FinCast consistently outperforms state-of-the-art methods, achieving reductions in forecasting error by an average of \textbf{20\%} and \textbf{23\%} respectively.
\end{itemize}

\section{Related Work}





Traditional financial time-series forecasting has historically relied on statistical models such as ARIMA~\cite{arima_2014}, GARCH~\cite{garch_stock}, and other domain-specific techniques~\cite{sampling_algo_stock_stats_cikm}. They often fall short in capturing nonlinear dynamics and abrupt domain shifts.

The emergence of deep learning introduced recurrent neural networks~\cite{medsker2001recurrent}, particularly Long Short-Term Memory~\cite{gers2000learning} architectures, as popular tools for modeling temporal dependencies in financial time series~\cite{lstm_stock_cikm24, lstm_2_cikm, cikm21_lstm_ts}. While these models can learn short- and medium-term dependencies, they tend to struggle with long-range correlations and suffer from vanishing gradients~\cite{lstm_cikm21_t}.

To account for inter-series relationships and interpretability, graph-based models have gained traction~\cite{graph_cikm_stock, graph_3_cikm, graph_ts_cikm20}. By representing stocks as nodes and their dependencies as edges, these methods incorporate relational inductive biases via explicit or learned topologies. Recent work has attempted to integrate graph-based models with sequence or variational frameworks~\cite{graph_vae_cikm24, graph_cikm_2}.

Transformer architectures~\cite{transformer_2017}, originally developed for NLP~\cite{achiam2023gpt} and later adapted to vision tasks~\cite{vit}, have shown promise in time-series forecasting~\cite{autoformer, informer, PatchTST} and financial time-series forecasting~\cite{PCIE_stock}. However, standard Transformers are computationally expensive and require architectural adaptations~\cite{AAAI_dlinear} to manage the irregularities and non-stationarity prevalent in financial data.

More recently, diffusion models have been proposed for financial time-series modeling~\cite{d-va, spatio_diff_stock_cikm24}. These models integrate diffusion processes with the latent representation learning of VAEs, allowing them to model uncertainty and complex temporal distributions. 

Other approaches—such as Bayesian models~\cite{martin2024bayesian} and reinforcement learning frameworks~\cite{RL_stock_cikm22}—have also been explored in financial contexts. These are often effective for niche tasks like trading policy learning or anomaly detection but face challenges in scalability and generalization.

Recent advances in large-scale models such as GPT-4~\cite{achiam2023gpt}, Claude, Gemini~\cite{team2023gemini}, and LLaMA~\cite{touvron2023llama} have demonstrated that combining architectural modularity, efficient routing~\cite{moe_gshard, stmoe}, and diverse pretraining can yield models with strong generalization across domains and tasks. 
Inspired by these advancements, foundation models for generic time series have recently been proposed. TimesFM~\cite{Timesfm}, TimesMoe~\cite{timemoe2024}, and Chronos-T5~\cite{chronos} are decoder-only transformer models pretrained on diverse timeseries datasets, demonstrating strong zero-shot capabilities. Yet, their design does not specifically address the idiosyncrasies of financial data, such as volatility, noise, and pattern shift. This motivates the development of \textbf{FinCast}, the first billion-parameter foundation model built explicitly for financial time-series forecasting.

\section{Methods}

\subsection{Problem Formulation}

We consider a financial time series $X_{1:L}= (x_1, \dots, x_{L}) \in \mathbb{R}^{L}$, where each $x_l \in \mathbb{R}$ is a scalar observation at time $l$. 
For any input context length $L \ge 1$ and forecast horizon $H \ge 1$, we define:
\begin{align}
X_{1:L}              &= [\,x_1,\dots,x_L\,] \;\in\;\mathbb{R}^{L},\\
X_{L+1:L+H}          &= [\,x_{L+1},\dots,x_{L+H}\,] \;\in\;\mathbb{R}^{H}.
\end{align}

Our goal is to learn a mapping $f_{\theta}:\mathbb{R}^{L} \to \mathbb{R}^{H}$ such that
\begin{equation}
\hat{X}_{L+1:L+H}
=
f_{\theta}(X_{1:L})\,,
\end{equation}

\noindent where $\hat{\mathbf{X}}_{L+1:L+H}$ denotes the forecast future values. Unlike conventional models that require fixed $L$ and $H$, FinCast supports arbitrary context length $L$ and horizon $H$ at inference time without changing the architecture. To handle variable feature dimension $c$, we adopt a channel‐independence mechanism~\cite{PatchTST}, applying the same mapping to each coordinate series:
\begin{equation}
\hat{x}^i_{L+1:L+H}
=
f_{\theta}\bigl(x^i_1,\dots,x^i_L\bigr),
\quad
i=1,\dots,c.
\end{equation}

\begin{figure*}[t]
\centering
\includegraphics[width=\linewidth]{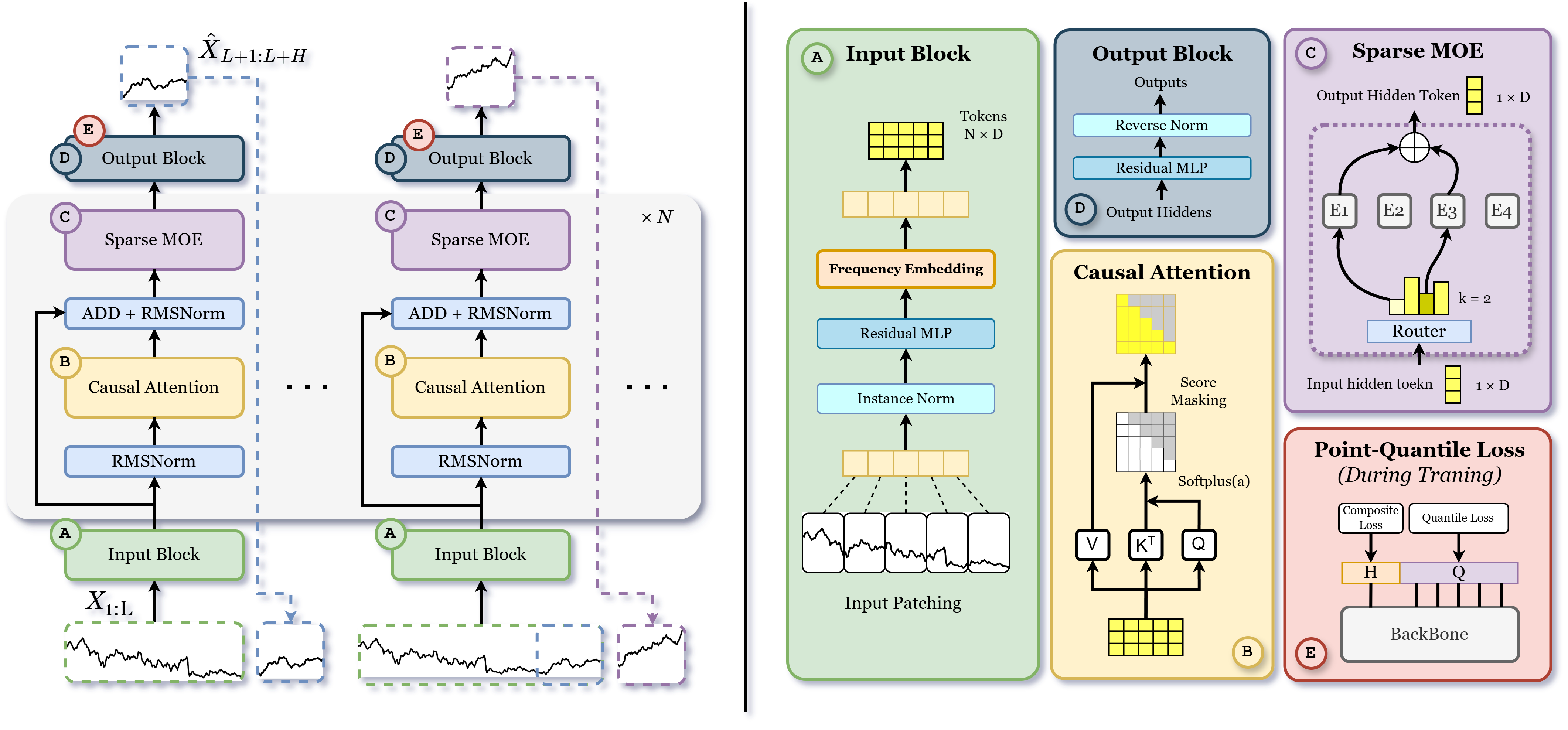}
\caption{FinCast Model Architecture. \textcolor{red}{\textcircled{\textcolor{black}{A}}}: Input preprocessing, tokenization, and applies Learnable Frequency Embedding. \textcolor{red}{\textcircled{\textcolor{black}{B}}}: Causal Attention, masking future attention scores. \textcolor{red}{\textcircled{\textcolor{black}{C}}}: Sparse MOE, activation based on each token. \textcolor{red}{\textcircled{\textcolor{black}{D}}}: Process decoder outputs and reverse norm. \textcolor{red}{\textcircled{\textcolor{black}{E}}}: PQ-Loss jointly optimizes output head.}
\Description{This fig shows the main architecture of the model}
\label{fig:FinCast}
\end{figure*}

\subsection{Model Architecture Overview}

FinCast is a decoder-only transformer architecture designed for financial time-series forecasting, illustrated in Figure~\ref {fig:FinCast}. It integrates our three key technical contributions: a token-level sparse Mixture-of-Experts to enable specialization across domains (Figure~\ref {fig:FinCast} Part C); learnable frequency embeddings to facilitate capturing resolution-specific temporal patterns (Figure~\ref {fig:FinCast} Part A); and a point-quantile loss that jointly optimizes accuracy and probabilistic estimates (Figure~\ref {fig:FinCast} Part E). 

The model consists of three principal components:

\begin{enumerate}
    \item \textbf{Input Tokenization Block:} The input time series is first normalized using instance normalization, then mapped into latent representations through residual MLP. Our learnable frequency embeddings are then injected to encode temporal resolution and periodicity.

    \item \textbf{Decoder MOE backbone:} A stack of Transformer decoder blocks with causal masking processes the latent tokens. FinCast employs a token-based sparse Mixture of Experts (MoE) mechanism, dynamically selecting experts per token.

    \item \textbf{Output Block:} The final hidden states are mapped to forecast outputs via residual MLP, followed by denormalization. The model is trained with point-quantile loss to jointly optimize point accuracy and probabilistic estimates.
\end{enumerate}

\subsection{Input Tokenization Block}

The input tokenization block transforms raw time series into patch-level tokens suitable for Transformer-based modeling. Given an input sequence $X \in \mathbb{R}^{B \times L}$, where $B$ is the batch size, $L$ is the sequence length, the sequence is first segmented into $N = \lfloor L / P \rfloor$ non-overlapping patches of length $P$, resulting in $X \in \mathbb{R}^{B \times N \times P}$.
\\
\textbf{Instance Normalization} Each input is then instance-normalized to ensure scale-invariant representations. For each input $X_{n, p}$, the normalization is given by:

\begin{equation}
\tilde{X}_{n,p} = \frac{X_{n,p} - \mu_{n}}{\sigma_{n}},   \quad
\mu_{n} = \frac{1}{P}\sum_{p=1}^{P}X_{n,p}
\end{equation}
\begin{equation}
\sigma_{n} = \sqrt{\frac{1}{P}\sum_{p=1}^{P}(X_{n,p}-\mu_{n})^2}.
\end{equation}

where $\tilde{X}$ denotes the normalized sequence, and $\mu, \sigma$ are the mean and standard deviation. During training, a binary mask $m_{n} \in \{0,1\}$ is used to mask part of the input, normalization is applied only to non-masked elements. The normalization parameters $\mu, \sigma$ are stored for inverse transformation during the Residual Output Block. Instance normalization offers three key advantages: (1) it removes scale bias, allowing the model to focus on dynamics and temporal structure; (2) it enhances robustness across financial instruments with varying magnitudes—crucial for a general-purpose financial foundation model; and (3) as a form of z-score normalization, it preserves the relative shape of the series in a lossless and reversible manner.
\\
\textbf{Input Residual Block} We then use a residual block which is a MLP with one hidden layer and a skip connection, similar to \cite{Timesfm}. 
The final model input is obtained by applying a linear projection to the concatenated vector:
\begin{equation}
h_{input} = \text{InputResidualBlock}((1 - M_{n}) \odot  \tilde{X}_{n} ) \in \mathbb{R}^{D_{\text{model}}}
\end{equation}
This produces a sequence of input tokens $h_{input} \in \mathbb{R}^{B \times N \times D_{\text{model}}}$, which are fed into the decoder-MOE backbone for forecasting.
\\

\textbf{Frequency Embedding} To support generalization across diverse temporal resolutions (e.g., minute-level, hourly, daily), FinCast employs a learnable frequency embedding mechanism. Each input sequence is assigned a discrete frequency index $f \in \mathbb{Z}$, which is used to retrieve a learnable embedding vector. After the residual MLP block, this vector is uniformly added to all $h_{input}$ in the sequence:
\begin{equation}
h_{\text{input}} = h_{\text{input}} + \text{Embed}_{\text{freq}}(f),
\end{equation}
where $\text{Embed}_{\text{freq}}: \mathbb{Z} \to \mathbb{R}^{D_{\text{model}}}$ is a learnable embedding function parameterized by the model. This component serves as an inductive bias, allowing the model to condition its internal representations on the temporal resolution of the input. By explicitly encoding frequency information, the model can more effectively learn resolution-specific patterns, enhancing its adaptability and forecast accuracy across diverse financial domains.

\subsection{Decoder MOE backbone}
\textbf{RMSNorm}
Formally, given an input token sequence $h_{input} \in \mathbb{R}^{B \times N \times D_{\text{model}}}$, the RMSNorm operation computes:
\begin{equation}
\text{RMSNorm}(h) = \gamma \cdot \frac{h}{\sqrt{\frac{1}{N} \sum_{i=1}^{N} h_i^2 + \epsilon}},
\end{equation}
where $\gamma \in \mathbb{R}^D$ is a learnable scale parameter and $\epsilon$ is a small constant for numerical stability. Unlike standard LayerNorm, RMSNorm omits mean subtraction, relying solely on the $\ell_2$ norm, which has been shown to be effective in large-scale pretraining~\cite{rmsnorm}.
\\

\textbf{Causal Self-Attention.} 
 Causal Self-Attention ensures autoregressive consistency in forecasting, where each token attends only to its current and past positions. Such masking is critical for financial time series forecasting to prevent information leakage from the future. Causal attention accommodates variable-length inputs and supports flexible forecast horizons. 
Let the normalized hidden states be denoted as $h_{\text{norm}} \in \mathbb{R}^{B \times N \times D_{\text{model}}}$. These are projected into query, key, and value tensors via a single linear transformation:
\begin{equation}
[Q, K, V] = h_{\text{norm}} W_{qkv},
\end{equation}
where $W_{qkv} \in \mathbb{R}^{D_{\text{model}} \times (H \cdot d_q + H \cdot d_k + H \cdot d_v)}$, with $H$ denoting the number of attention heads and $d_q = d_k = d_v$ the dimensionality per head. The projected tensors are reshaped to $Q, K, V \in \mathbb{R}^{B \times H \times N \times d_q}$.

Each query vector undergoes per-dimension reweighting:
\begin{equation}
Q' = Q \odot \left( \frac{\log_2 e}{\sqrt{d_q}} \cdot \text{softplus}(\alpha) \right),
\end{equation}
where $\alpha \in \mathbb{R}^{d_q}$ is a learned parameter vector shared across all heads, and $\odot$ denotes element-wise multiplication. This allows the model to adaptively scale each feature dimension within the attention computation \cite{Timesfm}.

To enforce autoregressive behavior and handle variable-length sequences, we apply an attention mask $M$, where masked entries are set to large negative values~\cite{Timesfm}. The attention logits are then computed as:
\begin{equation}
\text{Scores} = \text{softmax} \left( Q' K^\top + M \right),
\end{equation}
where the dot product is computed over the last dimension of $Q'$ and $K$.

The output of attention is a weighted combination of values:
\begin{equation}
\text{Attn}(h) = W_o \cdot (\text{Scores} \cdot V),
\end{equation}
where $W_o \in \mathbb{R}^{H \cdot d_v \times D_{\text{model}}}$ projects the concatenated heads back into the model dimension.


\textbf{Sparse Mixture-of-Experts.}
Following self-attention, the decoder block routes the residual-enhanced hidden states through a token-level Sparse Mixture-of-Experts (MoE) layer to increase representational capacity while maintaining computational efficiency. Each token is routed to its top-$k$ most suitable experts via a learned gating mechanism, enabling dynamic specialization across tokens. This design allows individual experts to capture distinct patterns and distributions commonly observed in financial time series, such as volatility bursts, seasonal shifts, and abrupt trend changes.
Let $h = h_{\text{res}} + \text{Attn}(h)$ denote the post-attention residual state. 
This is first normalized via RMSNorm, then passed into the MoE block.
For the token-level sparse gating mechanism with top-$k$ routing, where each token is routed to the $k$ most relevant experts based on a learned gating network. The dispatch tensor determines expert assignments, and token-expert interactions are aggregated via weighted combinations of expert outputs. Each expert consists of a lightweight two-layer MLP with residual connections.

Formally, each token $h_n \in \mathbb{R}^D$ for $n = \{1, \dots, N\}$ is routed to its top-$k$ experts via a learned gating mechanism. The gating logits are computed as:

\begin{equation}
s_{i,n} = \text{Softmax}_i(W_{\text{gate}} h_n),
\end{equation}
where $W_{\text{gate}} \in \mathbb{R}^{D \times E}$ projects the token to $E$ expert scores. Routing is sparse: only the top-$k$ scores are retained,
\begin{equation}
g_{i,n} = 
\begin{cases}
s_{i,n}, & \text{if } i \in \operatorname{Top}\text{-}k\left( \{s_{j,n}\}_{j=1}^{E} \right) \\
0, & \text{otherwise},
\end{cases}
\end{equation}
and the expert outputs are aggregated as:
\begin{equation}
\text{MoE}(h_n) = \sum_{i=1}^{E} g_{i,n} \cdot \text{MLP}_i(h_n),
\end{equation}
where $\text{MLP}_i$ denotes the $i$-th expert. The final token output is:
\begin{equation}
h'_n = h_n + \text{MoE}(\text{RMSNorm}(h_n)).
\end{equation}

The MoE layer improves robustness and expressivity by enabling specialization across diverse patterns. This design isolates noise and distributional shifts to specific experts, reducing interference in shared representations. 


\subsection{Output Block}


The final hidden states produced by the decoder are passed through the \textit{Residual Output Block}, which generates the forecast. Specifically, each token representation $h'_n \in \mathbb{R}^{D_{\text{model}}}$ is mapped to the output space via a residual feedforward block:
\begin{equation}
\hat{y}_n = \text{ResidualMLP}(h'_n) \in \mathbb{R}^{H},
\end{equation}
where $H$ is the forecast horizon length. The projection is implemented via a two-layer MLP with an intermediate nonlinearity and a residual connection, enhancing the output's capacity while preserving stable gradients. The sequence of outputs is then reshaped to form the tensor $\hat{Y} \in \mathbb{R}^{B \times N \times H}$.

To ensure consistency with the original data scale, FinCast performs an inverse normalization using the stored patch-level statistics $\mu, \sigma$ from the input tokenization phase:
\begin{equation}
\hat{Y}_{n,:} = \hat{y}_{n,:} \cdot \sigma_{n} + \mu_{n}.
\end{equation}
This rescaling is a lossless inverse normalization, restoring the original scale and ensuring forecasts are both accurate and directly comparable to raw inputs, critical in financial contexts where magnitude and scale semantics must be preserved across instruments and regimes.

\subsection{Point Quantile Loss}

A central contribution of our method is integrating a quantile-based loss as an auxiliary objective (Figure~\ref {fig:FinCast} Part E), which mitigates forecast collapse and enhances distributional robustness. The loss function is designed to enforce accurate, robust, and trend-consistent multi-step forecasts while promoting diversity and stability in the MOE block.  The total loss is a weighted sum of four components:
\begin{equation}
\mathcal{L}_{\text{total}} = \mathcal{L}_{\text{point}} + \mathcal{L}_{\text{trend}} +  \mathcal{L}_{\text{quantile}} +  \mathcal{L}_{\text{MOE}},
\end{equation}
where each $\lambda$ controls the relative contribution of its corresponding term.

\paragraph{Quantile Loss}
A key contribution in our loss design is the incorporation of a probability forecast objective by using quantile loss:
\begin{equation}
\mathcal{L}_{\text{quantile}} = \lambda_{\text{quantile}} \sum_{q \in \mathcal{Q}} \frac{1}{H} \sum_{t=1}^H 
\left\{
\begin{array}{ll}
q \cdot (y_t - \hat{y}^q_t), & \text{if } y_t \geq \hat{y}^q_t \\
(1 - q) \cdot (\hat{y}^q_t - y_t), & \text{otherwise}
\end{array}
\right.,
\end{equation}
where $\hat{y}^q_t$ denotes the $q$-th quantile forecast and $\mathcal{Q}$ is a set of quantiles (e.g. deciles). The quantile loss shapes the internal representations and promotes diversity in the learned distribution. It explicitly encourages the model to represent distributional asymmetries and capture forecast uncertainty. This design mitigates forecast collapse as shown by some model in figure~\ref{fig:zs_vis}, where models trained solely with MSE-based losses tend to regress toward the mean\cite{informer, AAAI_dlinear}.

\paragraph{Huber Point Loss}
The point forecast objective is a Huber loss\cite{huberloss2021} applied to the forecast mean $\hat{y} \in \mathbb{R}^H$:
\begin{equation}
\mathcal{L}_{\text{point}} = 
\frac{1}{H} \sum_{t=1}^H 
\begin{cases}
\frac{1}{2} (\hat{y}_t - y_t)^2, & \text{if } |\hat{y}_t - y_t| \leq \delta \\
\delta \cdot (|\hat{y}_t - y_t| - \frac{1}{2} \delta), & \text{otherwise}
\end{cases},
\end{equation}
which blends the benefits of MSE and MAE to preserve sensitivity for small errors while maintaining robustness to large deviations—particularly useful in high-noise environments.

\paragraph{Trend Consistency Loss.}
To align the local dynamics of forecast and actual series, we introduce a Trend Consistency Loss on first-order temporal differences: 
\begin{equation}
\mathcal{L}_{\text{trend}} = \lambda_{\text{trend}} 
\frac{1}{H-1} \sum_{t=2}^{H} 
\left( (\hat{y}_t - \hat{y}_{t-1}) - (y_t - y_{t-1}) \right)^2.
\end{equation}
This encourages the preservation of temporal trends and directional shifts—an essential feature for financial forecasting applications.

\paragraph{Auxiliary Expert Regularization.}
 
The auxiliary expert regularization loss includes a balance loss and a router z-loss:
\[
\mathcal{L}_{\text{MOE}} = \lambda_{\text{MOE}}( \mathcal{L}_{\text{balance}} +  \mathcal{L}_{\text{router-z}}),
\]
where:
\[
\mathcal{L}_{\text{balance}} = E \cdot \sum_k \bar{p}_k \bar{f}_k, \quad
\mathcal{L}_{\text{router-z}} = \mathbb{E}_{b,n} \left( \log \sum_k \exp(s_{b,n,k}) \right)^2.
\]
Here, $E$ is the number of experts, $B$ the batch size, $N$ the number of tokens, $s_{b,n,k}$ the gating logits, $\bar{f}_k$ the average assignment fraction, and $\bar{p}_k$ the mean gating probability for expert $k$.
$\mathcal{L}_{\text{balance}}$ promotes balance expert usage and $\mathcal{L}_{\text{router-z}}$ penalizes excessive entropy in the gating mechanism. This mitigates expert collapse and encourages specialization.

Overall, by aligning point forecasts with quantile-based uncertainty estimates, it enables the model to capture both central tendencies and tail risks.

\subsection{Model Training and Inference}

\subsubsection{Pretraining Dataset}

Training robust and generalizable foundation models for financial time series demands access to large-scale, high-quality and diverse datasets. We create a comprehensive pretraining dataset with \textbf{20 billion} time points across multiple financial and non-financial domains, encompassing a wide range of temporal frequencies from seconds to months. Table~\ref{tab:dataset-stats} summarizes the key statistics of our dataset.

The financial subset covers cryptocurrency, forex, futures, stocks, and macroeconomic indicators, each characterized by heterogeneous sampling rates and diverse structural dynamics. All financial data is obtained through publicly accessible interfaces and APIs. For the non-financial portion, we incorporate miscellaneous (Others) datasets sourced from \cite{climate_datasets}\cite{Timesfm}\cite{Monash_Dataset}\cite{timemoe2024} to facilitate the training of the model since high-quality financial data is scarce.
In total, the dataset comprises 2.4 million time series and more than 20 billion time points. We apply a rigorous data-cleaning pipeline to ensure training stability by removing invalid data, extreme outliers, and temporal inconsistencies.


\begin{table}[ht]
  \centering
  \caption{Statistics of the pretraining dataset}
  \label{tab:dataset-stats}
  \resizebox{\linewidth}{!}{%
    \begin{tabular}{lrrrrrr}
      \toprule
      \textbf{Domain} & \textbf{Crypto} & \textbf{Forex} & \textbf{Future} & \textbf{Stock} & \textbf{Econ} &  \textbf{Others} \\
      \midrule
      \# Time Series & 91{,}280 & 64{,}720 & 47{,}304 & 565{,}548 & 37{,}730 & 1{,}510{,}863 \\
      \# Time Points & 1.78B & 3.27B & 1.71B & 9.1B & 4.1M &  4.61B \\
      Percentage (\%) & 8.69\% & 15.96\% & 8.36\% & 44.49\% & 0.02\% & 22.48\% \\
      \bottomrule
    \end{tabular}
  }
\end{table}

\subsubsection{Training Details}

FinCast is a decoder-only, sparse Mixture-of-Experts (MoE) transformer with 1 billion parameters. For each sparse-MOE layer, it has 4 experts with a top-$k{=}2$ routing. Its design is motivated by scaling law insights~\cite{scalinglawsneurallanguage,trainingcomputeoptimallargelanguage}, which highlight the importance of model capacity when matched with sufficient data.

The model is trained with variable sequence lengths. The maximum training context length is 1024 for high-frequency series (e.g., seconds to daily). For coarser frequencies such as weekly to monthly, the training context is reduced to 256. 
We use a masking ratio of 15\% for our input patch, similar to~\cite{Timesfm}. Without masking, the model tends to generalize only to context lengths that are multiples of the input patch length.

FinCast undergoes 147{,}152 training steps, with each step processing approximately 5.2 million time points. Optimization is performed using the AdamW optimizer with a learning rate of $0.0002$ and a weight decay of 0.05. We train with a global batch size of 8192, 1024 per GPU across 8 NVIDIA H200 GPUs. For inter-GPU communication, we use nccl as the backend and implement distributed training with \texttt{DistributedDataParallel} from \texttt{torch.nn.parallel} and \texttt{torch.distributed}.

The learning rate schedule consists of a linear warmup over the first 5\% of training steps, followed by a 30\% stable plateau and a cosine decay to 10\% of the peak learning rate. All model weights are maintained in \texttt{Float32}, while training is executed with TF32 tensor cores and precision set to \texttt{high} to ensure numerical robustness without compromising throughput.

\subsubsection{Inference Procedure}
At inference time, FinCast operates in an auto-regressive decoding mode as shown in Figure~\ref{fig:FinCast}. It generates forecasts iteratively in patch-wise segments, with the output of each step appended to the end of the input for subsequent decoding. This patch-wise decoding continues until the desired forecast horizon is reached. Formally, for an input $X_{1:L}$, the model iteratively forecasts patches $\hat{X}_{L+1:L+H}, \hat{X}_{L+H+1:L+2H}, \ldots$ until the full horizon $H_{full}$ is covered. The final outputs consist of both the point forecast $\hat{X}_{L+1:L+H_{full}}$. Despite its scale, FinCast remains inference-efficient, capable of inferencing under full precision on a 8GB consumer-grade GPU as shown in figure~\ref{fig:speed_performance}.

\section{EXPERIMENTS}

We evaluate FinCast across two comprehensive forecasting benchmarks: \emph{Comparison to Zero-Shot Methods} and \emph{Comparison to Supervised Methods}, to comprehensively evaluate the performance. We also conduct extensive qualitative analyses, illustrating how FinCast handles shifting patterns across domains and temporal resolutions.

To evaluate zero-shot performance, we introduce a benchmark dataset comprising 3,632 time series and over \textbf{4.38} million scalar time points. The dataset reflects core challenges of real-world financial forecasting, including non-stationarity, diverse domains, and differences in temporal resolution.
As no specialized financial foundation models are publicly available, we compare FinCast against state-of-the-art general-purpose time-series foundational models, including Google’s TimesFM\cite{Timesfm} (200M parameters and 500M parameters versions), Amazon’s Chronos-T5\cite{chronos} (small, base, and large variants) and TimesMOE's large version\cite{timemoe2024}, all of which include financial time series data in their pretraining datasets.

In the supervised forecasting setting, we adopt the standardized benchmark from~\cite{PCIE_stock} for fair comparison. We report results for both the base FinCast (without fine-tuning) and a fine-tuned variant, evaluating against SOTA supervised models including PCIE\cite{PCIE_stock}, PatchTST\cite{PatchTST}, D-Va\cite{d-va}, Autoformer\cite{autoformer}, and Informer\cite{informer}.

\subsection{Comparison to Zero-Shot Methods}


We evaluate our model on a comprehensive financial time series benchmark.  It comprises 3,632 series with over \textbf{4.38} million scalar time points in total. Drawn from diverse financial domains, including cryptocurrencies, foreign exchange, stocks, and futures at varying temporal resolutions ranging from minute to weekly. 
The benchmark dataset is \textbf{excluded} from the pretraining datasets to ensure a strict zero-shot setting.
In contrast, existing general-purpose time series models may benefit from inadvertent overlap between their pretraining datasets and our benchmark, potentially inflating their performance due to information leakage.
We consider three forecast horizons, $h \in {10, 30, 60}$, which are commonly used by institutional investors and financial regulators~\cite{d-va}. We choose the input sequence length for all models $L = 128$ for fair comparison, following the standard practice recommended in the respective studies~\cite{ACL_18, Timesfm} to maintain fairness and comparability.


As shown in Table~\ref{tab:zeroshot_result}, our model consistently outperforms existing state-of-the-art methods across all forecast horizons. On average, \textit{FinCast} achieves a \textbf{20\%} reduction in MSE and a \textbf{10\%} reduction in MAE. It ranks first on 23 and 25 out of 36 diverse datasets, respectively. The benchmark’s scale and diversity make overfitting unlikely, so strong performance reflects genuine ability to model temporal dynamics and structural patterns in financial time series.


\begin{table*}[htbp]
\caption{Zero Shot Performance, Lower MSE and MAE indicates better results. Best Results are \textbf{bold}, second best are \underline{underline}}
\label{tab:zeroshot_result}
\resizebox{\textwidth}{!}{%
\begin{tabular}{ll|cc|cc|cc|cc|cc|cc|cc}
\toprule
\textbf{Models} & & 
  \multicolumn{2}{c|}{\textbf{FinCast(Ours)}} &
  \multicolumn{2}{c|}{\textbf{TimesFM}$_{\mathbf{200M}}$} &
  \multicolumn{2}{c|}{\textbf{TimesFM}$_{\mathbf{500M}}$} &
  \multicolumn{2}{c|}{\textbf{Chronos}$_{\textbf{Small}}$} &
  \multicolumn{2}{c|}{\textbf{Chronos}$_{\textbf{Medium}}$} &
  \multicolumn{2}{c|}{\textbf{Chronos}$_{\textbf{Large}}$} &
  \multicolumn{2}{c}{\textbf{TimesMOE}$_{\textbf{Large}}$} \\
     Metrics &  & MSE & MAE & MSE & MAE & MSE & MAE & MSE & MAE & MSE & MAE & MSE & MAE & MSE & MAE \\
\midrule
crypto\_1min & 10 & \underline{0.0114} & \underline{0.0683} & 0.0122 & 0.0703 & 0.0127 & 0.0709 & \textbf{0.0111} & 0.0684 & 0.0115 & \textbf{0.0682} & 0.0123 & 0.0700 & 0.0123 & 0.0706 \\
 & 30 & \textbf{0.0401} & \textbf{0.1256} & \underline{0.0419} & \underline{0.1283} & 0.0491 & 0.1393 & 0.0424 & 0.1320 & 0.0439 & 0.1331 & 0.0469 & 0.1363 & 0.0446 & 0.1284 \\
 & 60 & \underline{0.0837} & \underline{0.1850} & 0.0885 & 0.1908 & 0.1141 & 0.2107 & 0.0883 & 0.1944 & 0.0919 & 0.1991 & 0.1112 & 0.2130 & \textbf{0.0836} & \textbf{0.1823} \\
crypto\_1hour & 10 & \textbf{0.0090} & \textbf{0.0604} & 0.0099 & 0.0636 & 0.0106 & 0.0654 & 0.0097 & 0.0626 & \underline{0.0095} & \underline{0.0624} & 0.0097 & 0.0630 & 0.0102 & 0.0646 \\
 & 30 & \textbf{0.0236} & \textbf{0.1027} & 0.0259 & 0.1100 & 0.0325 & 0.1188 & 0.0254 & \underline{0.1052} & \underline{0.0247} & 0.1054 & 0.0256 & 0.1064 & 0.0278 & 0.1133 \\
 & 60 & \textbf{0.0440} & \textbf{0.1430} & 0.0530 & 0.1614 & 0.0620 & 0.1679 & 0.0508 & 0.1497 & \underline{0.0505} & \underline{0.1483} & 0.0516 & 0.1504 & 0.0570 & 0.1659 \\
crypto\_1day & 10 & 0.0572 & 0.1165 & 0.0655 & 0.1248 & 0.0951 & 0.1353 & \underline{0.0546} & \textbf{0.1152} & \textbf{0.0536} & \underline{0.1153} & 0.0578 & 0.1187 & 0.0672 & 0.1277 \\
 & 30 & 0.1445 & \textbf{0.1889} & 0.1889 & 0.2158 & 0.2937 & 0.2592 & \underline{0.1436} & 0.1943 & \textbf{0.1385} & \underline{0.1932} & 0.1600 & 0.2003 & 0.1823 & 0.2219 \\
 & 60 & 0.2774 & \textbf{0.2749} & 0.3588 & 0.3226 & 0.5730 & 0.3971 & \underline{0.2630} & \underline{0.2780} & \textbf{0.2502} & 0.2788 & 0.3105 & 0.3053 & 0.3316 & 0.3178 \\
 \midrule
forex\_1min & 10 & \textbf{0.0336} & \textbf{0.1182} & 0.0363 & 0.1227 & 0.0392 & 0.1248 & 0.0358 & 0.1206 & 0.0360 & 0.1206 & 0.0357 & 0.1216 & \underline{0.0353} & \underline{0.1190} \\
 & 30 & \textbf{0.0855} & \textbf{0.1897} & 0.0931 & \underline{0.1962} & 0.1166 & 0.2160 & 0.0939 & 0.2013 & \underline{0.0929} & 0.2005 & 0.0939 & 0.2007 & 0.1131 & 0.2087 \\
 & 60 & \textbf{0.1830} & \textbf{0.2671} & 0.2055 & 0.2893 & 0.2396 & 0.3096 & \underline{0.1907} & \underline{0.2786} & 0.1933 & 0.2827 & 0.2023 & 0.2900 & 0.2335 & 0.3050 \\
forex\_1day & 10 & \underline{0.0318} & \textbf{0.1250} & 0.0339 & 0.1289 & 0.0375 & 0.1330 & 0.0337 & 0.1293 & 0.0336 & 0.1298 & 0.0330 & 0.1278 & \textbf{0.0316} & \underline{0.1254} \\
 & 30 & \underline{0.0859} & \underline{0.2119} & 0.0976 & 0.2233 & 0.1089 & 0.2411 & 0.0897 & 0.2218 & 0.0895 & 0.2222 & 0.0889 & 0.2171 & \textbf{0.0799} & \textbf{0.2079} \\
 & 60 & \underline{0.1436} & \textbf{0.2726} & 0.1695 & 0.2981 & 0.1639 & 0.3013 & 0.1533 & 0.2925 & 0.1552 & 0.2911 & 0.1438 & 0.2819 & \textbf{0.1423} & \underline{0.2732} \\
forex\_1wk & 10 & \textbf{0.2076} & \textbf{0.3058} & 0.2520 & 0.3419 & 0.2555 & 0.3371 & 0.2266 & 0.3203 & 0.2228 & 0.3120 & \underline{0.2162} & 0.3081 & 0.2182 & \underline{0.3075} \\
 & 30 & \underline{0.3765} & \underline{0.4349} & 0.6104 & 0.5661 & 0.5174 & 0.5181 & 0.3966 & 0.4459 & 0.3887 & 0.4350 & \textbf{0.3582} & \textbf{0.4185} & 0.4222 & 0.4522 \\
 & 60 & \textbf{0.6041} & 0.5688 & 1.2389 & 0.8170 & 1.0439 & 0.7292 & 0.6270 & \underline{0.5653} & 0.7215 & 0.6039 & \underline{0.6251} & \textbf{0.5636} & 0.6302 & 0.5673 \\
 \midrule
futures\_1min & 10 & 0.1838 & 0.1986 & \underline{0.1743} & 0.1911 & 0.1843 & \underline{0.1870} & 0.2123 & \textbf{0.1847} & 0.2266 & 0.1965 & 0.2261 & 0.1938 & \textbf{0.1606} & 0.1969 \\
 & 30 & \underline{0.2092} & \underline{0.2470} & 0.2184 & 0.2506 & 0.2388 & 0.2569 & 0.2486 & \textbf{0.2414} & 0.2605 & 0.2547 & 0.2706 & 0.2543 & \textbf{0.2035} & 0.2601 \\
 & 60 & \textbf{0.2495} & \textbf{0.2936} & \underline{0.2716} & 0.3110 & 0.3220 & 0.3333 & 0.2965 & \underline{0.2975} & 0.3012 & 0.3054 & 0.3253 & 0.3132 & 0.2729 & 0.3245 \\
futures\_1day & 10 & \textbf{0.0354} & \textbf{0.1193} & 0.0408 & 0.1260 & 0.0442 & 0.1312 & 0.0426 & 0.1262 & 0.0401 & 0.1249 & \underline{0.0397} & \underline{0.1247} & 0.0409 & 0.1256 \\
 & 30 & \textbf{0.0931} & \textbf{0.1999} & 0.1178 & 0.2199 & 0.1535 & 0.2396 & 0.1045 & 0.2083 & 0.1036 & \underline{0.2059} & \underline{0.1030} & 0.2086 & 0.1119 & 0.2148 \\
 & 60 & \textbf{0.2200} & \underline{0.2911} & 0.2646 & 0.3208 & 0.3278 & 0.3335 & \underline{0.2244} & \textbf{0.2892} & 0.2294 & 0.2929 & 0.2369 & 0.2948 & 0.2276 & 0.2911 \\
futures\_1wk & 10 & \underline{0.0948} & \underline{0.2290} & 0.1277 & 0.2565 & 0.1099 & 0.2461 & 0.1081 & 0.2400 & 0.1045 & 0.2369 & 0.1028 & 0.2345 & \textbf{0.0936} & \textbf{0.2229} \\
 & 30 & \textbf{0.1740} & \textbf{0.3106} & 0.4032 & 0.4548 & 0.2235 & 0.3489 & 0.2272 & 0.3409 & 0.2110 & 0.3355 & 0.2190 & 0.3346 & \underline{0.1964} & \underline{0.3166} \\
 & 60 & \textbf{0.1794} & \textbf{0.3140} & 0.9060 & 0.6800 & 0.4466 & 0.4744 & \underline{0.3312} & \underline{0.4091} & 0.3489 & 0.4260 & 0.3893 & 0.4625 & 0.3764 & 0.4489 \\
 \midrule
stock\_1min & 10 & \textbf{0.1241} & \textbf{0.2170} & 0.1531 & 0.2390 & 0.1444 & 0.2352 & 0.1356 & \underline{0.2268} & 0.1369 & 0.2278 & 0.1385 & 0.2295 & \underline{0.1346} & 0.2272 \\
 & 30 & \textbf{0.2851} & \textbf{0.3454} & 0.3625 & 0.3929 & 0.3773 & 0.4063 & 0.3082 & \underline{0.3645} & 0.3089 & 0.3648 & 0.3130 & 0.3694 & \underline{0.3048} & 0.3719 \\
 & 60 & \textbf{0.5179} & \textbf{0.4848} & 0.6668 & 0.5586 & 0.6956 & 0.5842 & 0.5512 & 0.5107 & 0.5449 & 0.5095 & 0.5506 & 0.5158 & \underline{0.5183} & \underline{0.5046} \\
stock\_1day & 10 & \textbf{0.0602} & \textbf{0.1488} & 0.0661 & 0.1558 & 0.0679 & 0.1581 & \underline{0.0632} & \underline{0.1527} & 0.0635 & 0.1532 & 0.0639 & 0.1537 & 0.0633 & 0.1530 \\
 & 30 & \textbf{0.1587} & \textbf{0.2479} & 0.1813 & 0.2661 & 0.1969 & 0.2782 & 0.1647 & 0.2558 & 0.1649 & 0.2567 & 0.1672 & 0.2579 & \underline{0.1621} & \underline{0.2556} \\
 & 60 & \underline{0.2887} & \underline{0.3440} & 0.3436 & 0.3750 & 0.3619 & 0.3887 & 0.2932 & 0.3550 & 0.2953 & 0.3571 & 0.2979 & 0.3586 & \textbf{0.2662} & \textbf{0.3412} \\
stock\_1wk & 10 & \textbf{0.1064} & \textbf{0.2125} & 0.1396 & 0.2408 & 0.1351 & 0.2359 & 0.1201 & 0.2231 & 0.1198 & \underline{0.2229} & 0.1205 & 0.2237 & \underline{0.1190} & 0.2231 \\
 & 30 & \textbf{0.2142} & \textbf{0.3056} & 0.3914 & 0.4123 & 0.3342 & 0.3839 & 0.2833 & 0.3578 & 0.2782 & 0.3542 & 0.2803 & 0.3561 & \underline{0.2759} & \underline{0.3541} \\
 & 60 & \textbf{0.2810} & \textbf{0.3606} & 0.7239 & 0.5758 & 0.5486 & 0.5150 & 0.4459 & 0.4725 & 0.4438 & 0.4696 & 0.4536 & 0.4751 & \underline{0.4368} & \underline{0.4636} \\
\midrule
Average &  & \textbf{0.1644} & \textbf{0.2397} & 0.2537 & 0.2888 & 0.2411 & 0.2836 & 0.1860 & \underline{0.2537} & 0.1886 & 0.2554 & 0.1911 & 0.2570 & \underline{0.1858} & 0.2571 \\
\bottomrule
\end{tabular}%
}
\end{table*}


\begin{table*}[ht]
\caption{Supervised Performance, Lower MSE and MAE indicates better results. Best Results are \textbf{bold}, second best are \underline{underline}}
\label{tab:id_result}
\centering
\resizebox{\textwidth}{!}{%
\begin{tabular}{ll|cc|cc|cc|cc|cc|cc|cc}

\toprule
\textbf{Models} & & 
  \multicolumn{2}{c|}{\textbf{FinCast\_finetune}} &
  \multicolumn{2}{c|}{\textbf{FinCast\_zeroshot}} &
  \multicolumn{2}{c|}{\textbf{PCIE}~\cite{PCIE_stock}} &
  \multicolumn{2}{c|}{\textbf{PatchTST}~\cite{PatchTST}} &
  \multicolumn{2}{c|}{\textbf{D-Va}~\cite{d-va}} &
  \multicolumn{2}{c|}{\textbf{Autoformer}~\cite{autoformer}} &
  \multicolumn{2}{c}{\textbf{Informer}~\cite{informer}} \\
Metrics & & MSE & MAE & MSE & MAE & MSE & MAE & MSE & MAE & MSE & MAE & MSE & MAE & MSE & MAE \\
\midrule
US\_71 & 10 & \textbf{0.0654} & \textbf{0.1732} & \underline{0.0675} & \underline{0.1766} & 0.0690 & 0.1784 & 0.0851 & 0.1903 & 0.2229 & 0.3338 & 0.1292 & 0.2584 & 0.1527 & 0.2904 \\
       & 20 & \textbf{0.1220} & \textbf{0.2368} & \underline{0.1271} & \underline{0.2451} & 0.1352 & 0.2554 & 0.1650 & 0.2985 & 0.2047 & 0.3193 & 0.2112 & 0.3261 & 0.3483 & 0.4271 \\
       & 40 & \textbf{0.2246} & \textbf{0.3271} & \underline{0.2361} & \underline{0.3403} & 0.2635 & 0.3618 & 0.2986 & 0.3987 & 0.3269 & 0.4240 & 0.3134 & 0.4103 & 0.3802 & 0.4613 \\
       & 60 & \textbf{0.2998} & \textbf{0.3793} & \underline{0.3129} & \underline{0.3943} & 0.3337 & 0.4156 & 0.3787 & 0.4496 & 0.4190 & 0.4895 & 0.3897 & 0.4593 & 0.4351 & 0.5010 \\
\midrule
US\_14L & 10 & \textbf{0.1454} & \textbf{0.2579} & 0.1509 & 0.2650 & \underline{0.1458} & \underline{0.2590} & 0.1655 & 0.2782 & 0.3472 & 0.4046 & 0.3009 & 0.3881 & 0.2573 & 0.3510 \\
        & 20 & \textbf{0.2730} & \textbf{0.3545} & \underline{0.2792} & 0.3643 & 0.2794 & \underline{0.3625} & 0.2942 & 0.3736 & 0.3893 & 0.4562 & 0.4543 & 0.4789 & 0.3285 & 0.3970 \\
        & 40 & \textbf{0.5016} & \textbf{0.4887} & \underline{0.5263} & \underline{0.5048} & 0.5570 & 0.5203 & 0.5705 & 0.5242 & 0.7245 & 0.6120 & 0.7498 & 0.6275 & 0.7037 & 0.6043 \\
        & 60 & \textbf{0.7454} & \textbf{0.5864} & \underline{0.7733} & \underline{0.6133} & 0.8251 & 0.6355 & 0.8488 & 0.6446 & 0.9461 & 0.7012 & 0.9885 & 0.7248 & 0.9257 & 0.6990 \\
\midrule
Average &  & \textbf{0.2971} & \textbf{0.3505} & \underline{0.3092} & \underline{0.3630} & 0.3261 & 0.3736 & 0.3508 & 0.3947 & 0.4476 & 0.4676 & 0.4421 & 0.4592 & 0.4414 & 0.4664 \\
\bottomrule
\end{tabular}%
}
\end{table*}

\subsection{Comparison to Supervised Methods}

We adopt two financial time series datasets from the PCIE benchmark: \textbf{US\_71} and \textbf{US\_14L}. The \textbf{US\_71} dataset consists of historical daily prices for 71 high-volume U.S. stocks, representing the top 6–9 stocks by market capitalization and trading volume across the nine major industry sectors. This construction follows established practices in prior stock forecasting literature~\cite{ACL_18,KDD17}. The data spans from 2016-01-04 to 2023-12-29.
The \textbf{US\_14L} dataset includes 14 large-cap, high-liquidity U.S. stocks, with daily historical prices collected over a longer period from 2005-01-04 to 2023-12-29. 
We partition each dataset into training, validation, and testing sets using a consistent 7:1:2 ratio across all models to ensure fair evaluation. Both datasets are \textbf{excluded} from the pretraining dataset of our model.

For supervised forecasting, we evaluate both the base (zero-shot) and finetuned versions of our model. The finetuned variant is trained on the respective training splits of the target datasets to assess performance under distributional alignment. Fine-tuning is performed with a lightweight and simple strategy: the model is trained for 1 epoch, with gradient updates restricted to the output block and the last 10\% of the sparse MoE layers. This setup evaluates the adaptability of the FinCast under minimal task-specific tuning.

According to Table~\ref{tab:id_result}, both the zero-shot and finetuned versions of our model surpass all existing state-of-the-art supervised models. Most notably, the zero-shot variant alone achieves a substantial performance gain, reducing MSE by \textbf{23\%} and MAE by \textbf{16\%} on average. The performance further improves with fine-tuning, yielding \textbf{26\%} and \textbf{19\%} reductions in MSE and MAE, respectively. These results underscore the robustness of our model, with the zero-shot variant alone outperforming all state-of-the-art supervised baselines, demonstrating its capacity to generalize effectively to unseen financial domains without task-specific adaptation.


\subsection{Ablation Study}

To quantify the individual contributions of our architectural and loss function design choices, we conduct a systematic ablation study on the zero-shot forecasting benchmark. Table~\ref{tab:ablation} reports the average MSE, MAE and performance degradation across the benchmark.

\begin{figure}[ht]
  \centering
  \includegraphics[width=\columnwidth]{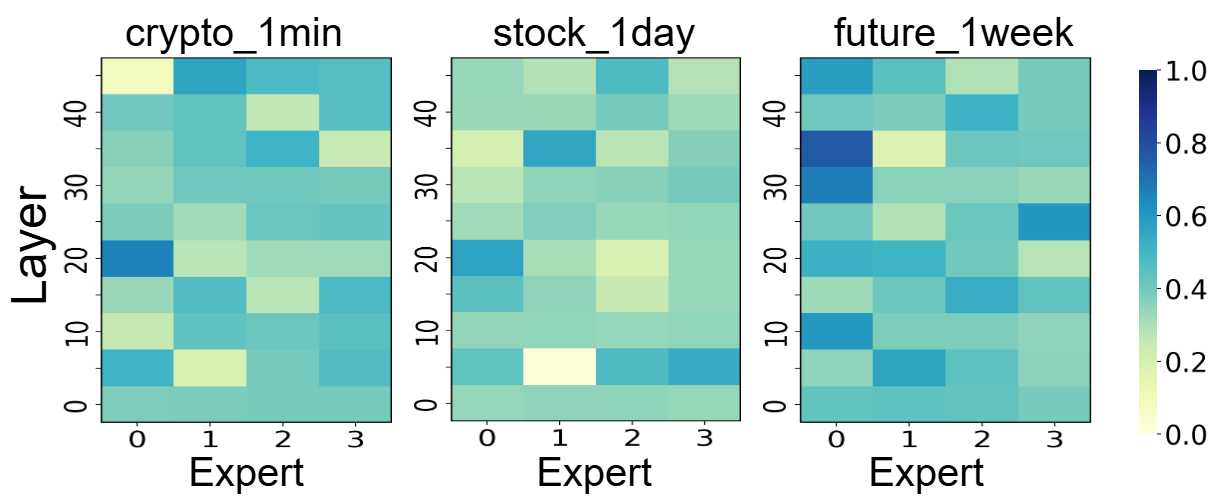}
  \caption{Expert activation patterns across datasets. Each expert specializes on domain-specific characteristics.}
  \Description{Expert activation patterns}
  \label{fig:expert_activation}
\end{figure}

\textbf{Sparse Mixture-of-Experts (MoE).} Replacing our token-level sparse MoE with a dense variant—where all experts are uniformly active—results in a substantial degradation of performance (\textbf{+9.32\%} MSE). This highlights the critical role of sparse, input-adaptive routing in promoting both generalization and specialization. As illustrated in Fig.~\ref{fig:expert_activation}, sparse gating enables distinct experts to specialize across financial domains and temporal resolutions, whereas dense routing induces homogenization and suppresses diversity.

\begin{figure}[ht]
  \centering
  \includegraphics[width=\columnwidth]{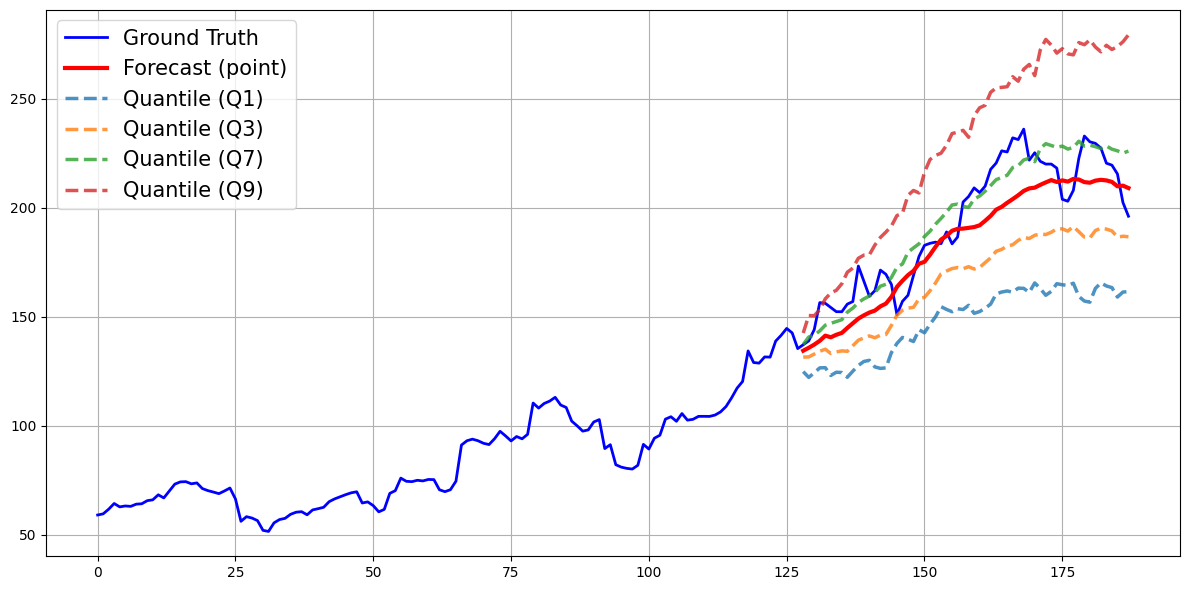}
  \caption{Point and Quantile Outputs During Training}
  \Description{PQ output visualization}
  \label{fig:PQL}
\end{figure}

\textbf{Point-Quantile Loss.} Training with a standard MSE loss instead of our proposed PQ-loss degrades performance by \textbf{7.62\%}. This confirms the advantage of PQ-loss in enhancing forecast robustness and preventing forecast collapse. Unlike an MSE loss, which tends to regress toward the mean~\cite{autoformer}. PQ-loss is especially robust under non-stationary conditions, where future distributions can shift unpredictably. As illustrated in Fig.~\ref{fig:PQL}, PQ-loss enables the model to capture distributional knowledge and uncertainty, crucial in the presence of pattern shifts.

\textbf{Frequency Embedding.} Excluding frequency embeddings causes a \textbf{4.38\%} performance degradation. This component serves as a critical inductive bias, allowing the model to condition on temporal resolution. Without frequency conditioning, the model is forced to infer temporal resolutions implicitly, which can lead to inconsistent behavior across different temporal resolutions. By using a learnable frequency embedding, FinCast explicitly encodes resolution information, enabling the model to adjust its internal representations according to the sampling rate. The ablation confirms that temporal resolution is a structural property that must be explicitly modeled for robust generalization.

\begin{table}[htbp]
  \caption{Ablation study results on MSE and MAE metrics. Lower is better.}
  \label{tab:ablation}
  \centering
    \resizebox{\linewidth}{!}{%
  \begin{tabular}{lccc}
    \toprule
    \textbf{Model Variant} & \textbf{MSE} & \textbf{MAE} & \textbf{ Degradation (\%)} \\
    \midrule
    \textbf{FinCast} & \textbf{0.1644} & \textbf{0.2397} & \textbf{-} \\
    \quad w/o sparse MOE & 0.1802 & 0.2617 & -9.32\% \\
    \quad w/o PQ-loss  & 0.1767 & 0.2582 & -7.62\% \\
    \quad w/o Freq Embedding & 0.1713 & 0.2505 & -4.38\% \\
    \bottomrule
  \end{tabular}
  }
\end{table}

\begin{figure*}[!t]
  \centering
  \includegraphics[width=\textwidth]{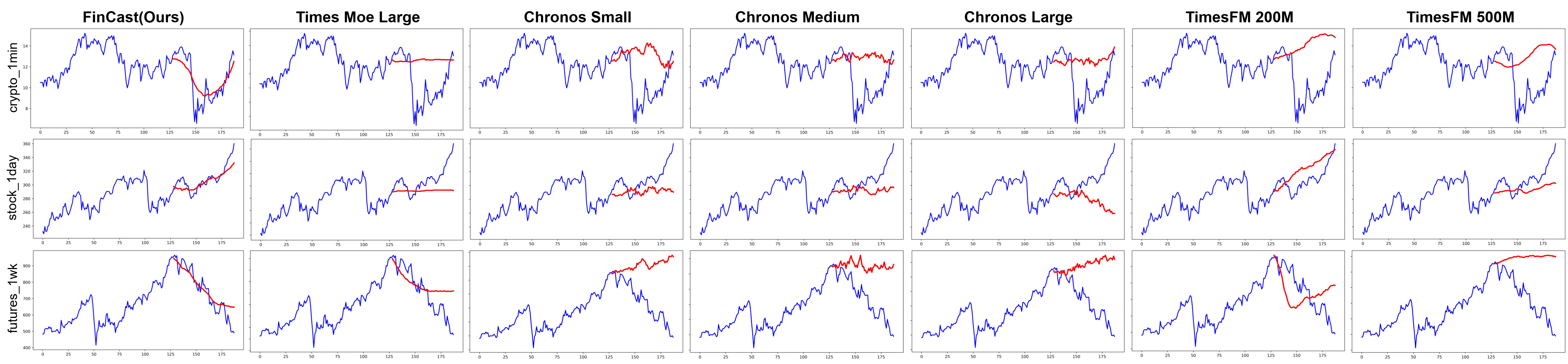}
  \caption{Zero shot forecasting examples from Zero Shot Forecast Benchmark, \textcolor{blue}{Blue} : Ground Truth, \textcolor{red}{Red} : Forecast}
  \Description{Qualitative examples to visualize}
  \label{fig:zs_vis}
\end{figure*}

\begin{figure*}[!t]
  \centering
  \includegraphics[width=\textwidth]{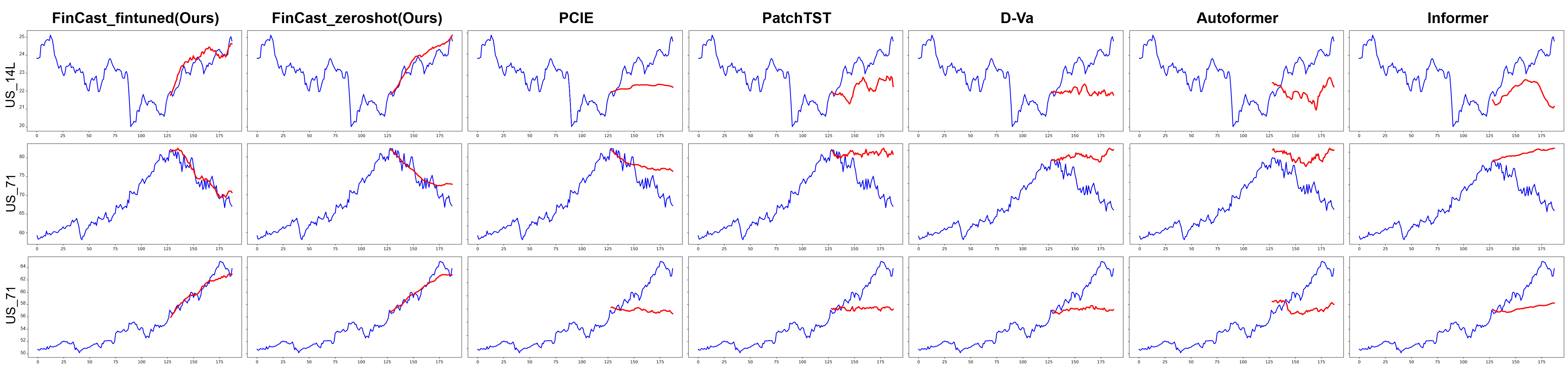}
  \caption{Supervised forecasting examples from Supervised Forecast Benchmark, \textcolor{blue}{Blue} : Ground Truth, \textcolor{red}{Red} : Forecast}
  \Description{Qualitative examples to visualize}
  \label{fig:sp_vis}
\end{figure*}

\subsection{Inference Speed Analysis}

\begin{figure}[ht]
  \centering
  \includegraphics[width=0.9\columnwidth]{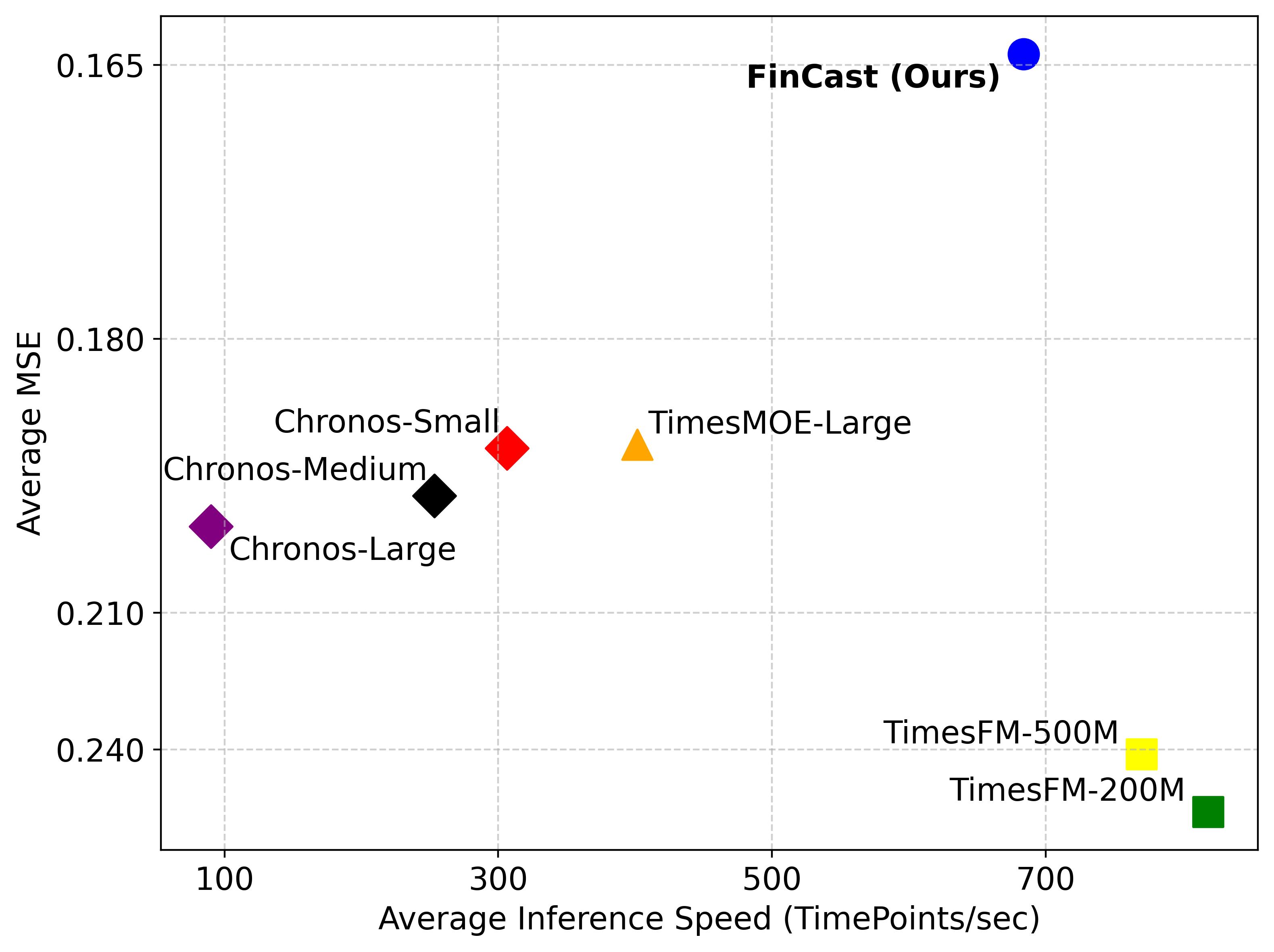}
  \caption{Inference Speed vs Performance}
  \Description{Inference quality visualization}
  \label{fig:speed_performance}
\end{figure}

Efficient inference is a critical requirement for deploying forecasting models in real-world financial settings, particularly in high-frequency trading, portfolio risk monitoring, and real-time market analytics~\cite{taylor2011asset}. These applications demand not only forecast accuracy but also minimal latency and hardware efficiency. As illustrated in Figure~\ref{fig:speed_performance}, FinCast achieves a favorable balance between inference speed and forecasting performance, significantly outperforming existing models along this trade-off frontier. It achieves up to 5× faster inference speed while outperforming all of the other generic time-series models in accuracy.

We report the average inference speed measured across benchmarks conducted in our zero-shot forecasting evaluations. Experiments were executed on a consumer-grade NVIDIA RTX 4060 GPU with 8GB of VRAM, which is a realistic proxy for deployment in production systems with constrained computational resources. 

FinCast’s inference efficiency derives from two key design choices. First, its token-level sparse Mixture-of-Experts (MoE) architecture activates only a subset of specialized experts per token, enabling conditional computation that significantly reduces inference cost without compromising capacity. Second, FinCast employs patch-wise tokenization rather than point-wise encoding~\cite{timemoe2024}, effectively reducing sequence length and thus lowering the computational burden of autoregressive decoding. 


\subsection{Qualitative Results}


Figure~\ref{fig:zs_vis} presents qualitative examples from the zero-shot dataset, which includes crypto\_1min, stock\_1day, and futures\_1wk, spanning diverse financial domains and temporal resolutions with non-stationary distributions. Most state-of-the-art models fail to generalize in these settings, some collapse to flat-line outputs due to only using MSE for optimization, while others struggle to capture the underlying pattern and distribution due to limited capacity. In contrast, FinCast demonstrates strong pattern sensitivity and trend awareness, accurately adapting to complex pattern shifts and diverse domains with different temporal resolutions.


Figure~\ref{fig:sp_vis} illustrates qualitative examples from the supervised dataset. These results highlight a fundamental limitation of supervised models, their tendency to regress toward the mean when faced with distributional uncertainty. In the final example, all baselines output flat-line forecasts due to a subtle but abrupt drop in the final input window. This behavior derives from their limited exposure to diverse patterns and distributions during training, leading them to default to conservative, low-variance outputs when uncertain. While such forecasts may not severely impact average error metrics, they are ineffective in practice, offering little beyond what simple statistical methods can produce, which completely defeats the point of using complex neural networks. This limitation also explains why many financial practitioners remain skeptical of supervised neural networks and often favor simpler statistical methods~\cite{sezer2020financial_ts}. Supervised models rely heavily on limited historical data and implicitly assume that future distributions will resemble those seen during training, an assumption rarely valid in real-world financial markets where various underlying pattern shifts.

\section{Conclusion and Future Works}

In summary, we introduced \textbf{FinCast}, the first foundation model tailored for financial time series forecasting. FinCast is designed to address the core challenges of non-stationarity, multi-domain diversity, and multi-temporal resolution, without requiring task-specific fine-tuning. 

Through extensive evaluation, FinCast achieves on average \textbf{20\% lower MSE} in zero-shot settings compared to existing SOTA methods. Qualitative analyses confirm that it avoids common failure modes such as flat-line outputs and mean reversion, instead producing trend-aware, high-fidelity forecasts. 

For future work, we aim to pretrain the model on larger and more diverse high-quality datasets.


Model weights, code can be found on: \url{https://github.com/vincent05r/FinCast-fts}

\clearpage

\section*{Generative AI Usage Statement}
The authors confirm that we did not use any generative AI tools (e.g. ChatGPT, Gemini, llama) during any stages of this research work. All aspects of the coding, research, writing, analysis, and figure preparation were performed solely by the authors without AI assistance.

\bibliographystyle{ACM-Reference-Format}
\bibliography{reference}


\end{document}